\documentclass[letterpaper]{article}
\usepackage[preprint]{aaai2027}
\usepackage[hyphens]{url}
\usepackage{graphicx}
\usepackage{natbib}
\usepackage{caption}
\usepackage{amsmath}
\usepackage{amssymb}
\usepackage{booktabs}
\usepackage{enumitem}
\usepackage[most]{tcolorbox}
\usepackage{tikz}
\usetikzlibrary{positioning,arrows.meta}
\usepackage[
    hidelinks,
    bookmarksnumbered=true,
    bookmarksopen=true
]{hyperref}
\definecolor{TitleBlack}{HTML}{222222}
\newtcolorbox{observationbox}[1][]{
    enhanced,
    colback=white,
    colframe=black,
    colbacktitle=TitleBlack,
    coltitle=white,
    fonttitle=\bfseries,
    title={#1},
    boxrule=0.6pt,
    arc=1.5pt,
    outer arc=1.5pt,
    left=7pt,
    right=7pt,
    top=4pt,
    bottom=4pt,
    before skip=4pt,
    after skip=4pt,
    titlerule=0pt
}

\newcommand{\model}{H\textsuperscript{+}~Embedding}
\newcommand{\modelmrl}{H\textsuperscript{+}~Embedding~MRL}
\newcommand{\vg}{\mathbf{g}}
\newcommand{\vh}{\mathbf{h}}
\newcommand{\vp}{\mathbf{p}}

\title{H\textsuperscript{+} Embedding: Harmonizing Global and Token-Level Retrieval with Context-Dependent Phrases}
\author{
    Shusen Zhang\equalcontrib,
    Junyi Hu\equalcontrib,
    Ye Feng,
    Ziteng Wang,
    Zhaoyuan Pan,
    Xiaojun Yuan,\\
    Jiangshou Hong,
    Guosheng Dong\corresponding,
    Xiangzhi Wang
}
\affiliations{
    Alibaba Health
}

\begin{document}
\maketitle

\begin{abstract}
Terminology-intensive retrieval, especially in medical settings, depends on preserving multi-word entities, abbreviations, numerical constraints, and compositional concepts. However, existing representations lie at two extremes: single-vector retrievers often over-compress local relevance signals, while token-level late interaction retains every tokenizer subword at substantial indexing, storage, and scoring cost. This mismatch raises a natural question: \emph{can context-dependent phrases provide a useful retrieval unit between global vectors and tokens?} We introduce \textbf{\model{}}, a unified multi-granularity retriever that predicts variable-length phrase partitions, preserves uncovered tokens as singletons, and applies importance-guided unit selection with weighted MaxSim interaction. Across 16 scientific, medical, and bilingual tasks, its phrase retrieval branch exceeds the global retrieval branch by 6.91 macro nDCG@10. It also nearly matches Token while using 13.7\% fewer document vectors and outperforms content-independent grouping rules under moderate vector budgets. Context-dependent phrase interaction therefore provides an intermediate quality-cost point between global compression and token-level interaction for practical retrieval systems.
\end{abstract}

\section{Introduction}
\label{sec:introduction}

\begin{figure*}[t]
\centering
\includegraphics[width=0.95\textwidth]{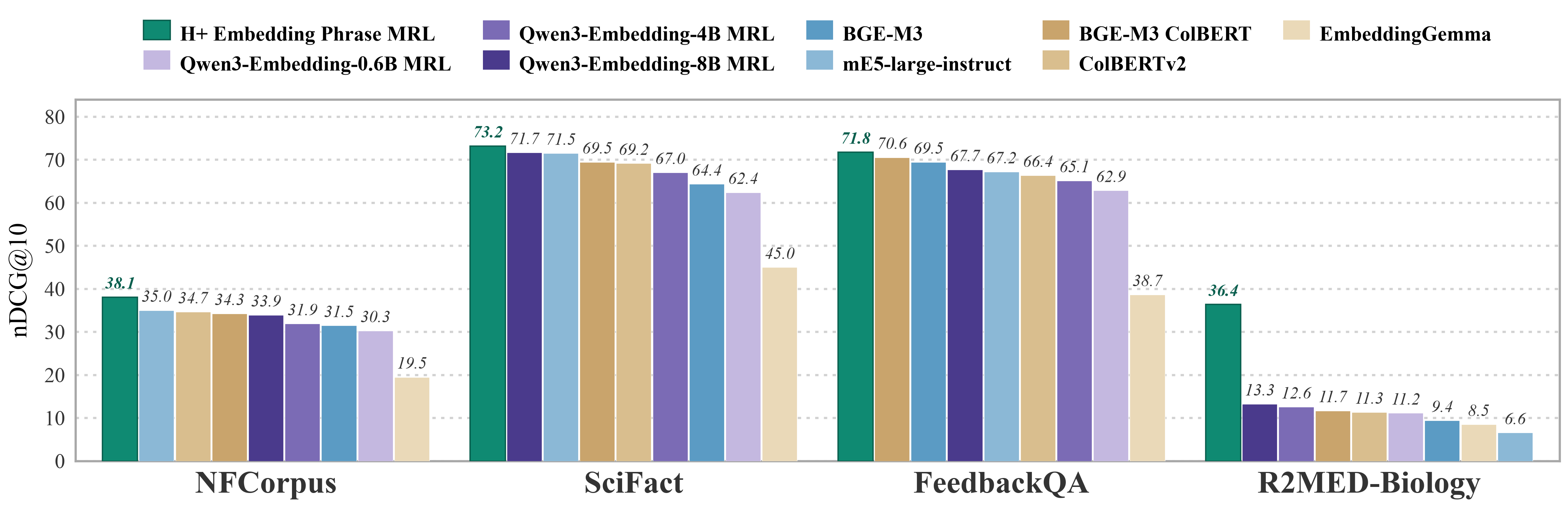}
\caption{Selected-task nDCG@10 under the CRF-only dimensional setting. Teal denotes \modelmrl{} Phrase (128-d, own Global top-1,000); dense baselines use full-corpus retrieval. Full results are provided in Appendix~\ref{sec:baseline_appendix}.}
\label{fig:sota_four_task}
\end{figure*}

Terminology-intensive retrieval is central to applications such as clinical evidence search, biomedical question answering, scientific literature review, and domain-specific knowledge search, where relevance often depends on preserving the internal structure of specialized expressions. Queries may combine multi-word entities, abbreviations, interventions, numerical constraints, and compositional conditions. An effective retriever must therefore capture not only the global intent of a query, but also the local structure through which specialized concepts are expressed and composed.

However, existing retrieval paradigms do not meet this requirement. Single-vector retrievers compress an entire query or document into one embedding, enabling efficient approximate nearest-neighbor search but potentially suppressing distinct local concepts \citep{reimers2019sbert,karpukhin2020dpr}. At the other extreme, token-level late-interaction models preserve fine-grained relevance signals through contextualized MaxSim, but treat every tokenizer subword as an independent retrieval unit and incur storage and scoring costs that grow with sequence length \citep{khattab2020colbert,santhanam2022colbertv2}. Learned sparse retrieval retains exact lexical signals, but does not resolve the underlying choice of semantic interaction granularity \citep{formal2021splade}. This mismatch raises a natural question: \emph{can context-dependent phrases provide a more appropriate retrieval unit between global vectors and individual tokens?}

\begin{observationbox}[The Retrieval Granularity Gap]
\[
\underbrace{\text{Global}}_{\text{one vector}}
\;\longrightarrow\;
\underbrace{\text{Phrase}}_{\text{context-dependent units}}
\;\longrightarrow\;
\underbrace{\text{Token}}_{\text{subwords}}
\]
\end{observationbox}

In this work, we define a \emph{retrieval phrase} as a contiguous, context-dependent span whose tokens should act jointly as one unit of local relevance. Unlike a tokenizer word or fixed $n$-gram, such a unit is determined by its meaning in context rather than by a predefined grouping rule. Learning these units under a finite vector budget introduces two coupled problems: the retriever must partition a text into local units and then determine which units should be retained. We therefore formulate retrieval-unit design as the joint problem of \emph{context-dependent partitioning} and \emph{budgeted unit selection}.

\begin{figure}[t]
    \centering
    \includegraphics[width=0.96\columnwidth]{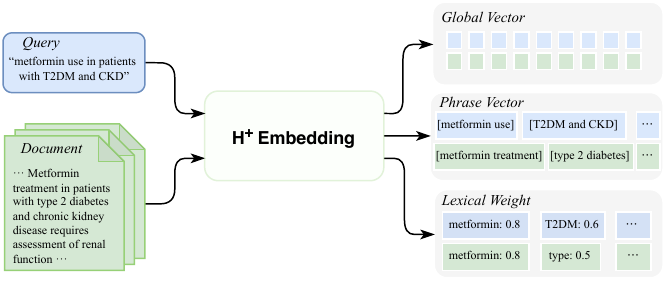}
    \caption{Overview of \model{}. A shared encoder produces global, phrase-level, and lexical representations, grouping medical concepts such as ``type 2 diabetes'' and ``chronic kidney disease'' into phrase-level retrieval units.}
    \label{fig:overview}
\end{figure}

To address this problem, we introduce \textbf{\model{}}, a unified multi-granularity retriever initialized from Qwen3-0.6B-Base \citep{yang2025qwen3}. As illustrated in Figure~\ref{fig:overview}, \model{} uses a shared bidirectional encoder to produce global, phrase-level, and lexical representations. To construct the intermediate retrieval units, it predicts context-dependent phrase boundaries with a Conditional Random Field (CRF), retains uncovered tokens as singleton units, and applies budgeted unit selection with aggregated token importance for weighted phrase-level MaxSim retrieval.

Extensive experiments establish the effectiveness, mechanism, and quality-cost trade-off of phrase interaction. Under candidate-constrained evaluation across 16 scientific, medical, and bilingual tasks, Phrase improves its Global ranking by 6.91 macro nDCG@10. Across 19 exact-budget tasks, Phrase exceeds Token throughout and surpasses Bigram and Random from $B=32$ onward. Boundary and weighting ablations identify complementary mechanisms, while deployed Phrase nearly matches Token on four tasks with 13.7\% fewer document vectors. Our contributions are threefold:
\begin{itemize}[leftmargin=1.3em, itemsep=1pt, topsep=2pt, parsep=0pt]
    \item We identify a granularity mismatch in terminology-intensive retrieval and formulate retrieval-unit design as the joint learning of context-dependent units and budgeted unit selection under practical vector constraints.

    \item We introduce \textbf{\model{}}, a unified global-phrase-lexical retriever with a shared encoder for multi-granularity retrieval research and applications.

    \item We show that context-dependent phrases improve retrieval quality under practical vector budgets and provide a favorable trade-off against token-level interaction.
\end{itemize}

\section{Problem Formulation}
\label{sec:problem}

Given a query $q$, a corpus $\mathcal{D}$, and local-vector budgets $(B_q,B_d)$, \model{} provides global and phrase retrieval branches together with an auxiliary lexical scoring view. Global retrieval uses one vector per document, while phrase retrieval stores up to $B_d$ local vectors per document and scores with weighted MaxSim over at most $B_q$ query units. For each retrieval branch $v\in\{G,P\}$,
\begin{equation}
    \mathcal{C}_{K_v}^{v}(q)
    =
    \operatorname{TopK}_{d\in\mathcal{D}}
    S_v(q,d).
    \label{eq:branch_retrieval}
\end{equation}
The two retrieval branches may be used independently. For hybrid
retrieval, their candidates are combined into
\begin{equation}
    \mathcal{C}(q)
    =
    \bigcup_{v\in\mathcal{V}}
    \mathcal{C}_{K_v}^{v}(q),
    \qquad
    \mathcal{V}\subseteq\{G,P\},
    \label{eq:candidate_union}
\end{equation}
and documents in the candidate union are ranked using
\begin{equation}
    S(q,d)
    =
    \alpha_G S_G(q,d)
    +
    \alpha_P S_P(q,d)
    +
    \alpha_L S_L(q,d),
    \label{eq:fusion}
\end{equation}
where the lexical score is evaluated over the retrieved candidates. The learning problem is to determine which tokens form joint phrase units and which units to retain under $(B_q,B_d)$. The weighted interaction rule then determines how strongly the retained query units contribute.

\section{H\textsuperscript{+}~Embedding}
\label{sec:method}
\begin{figure}[t]
    \centering
    \includegraphics[width=0.96\columnwidth]{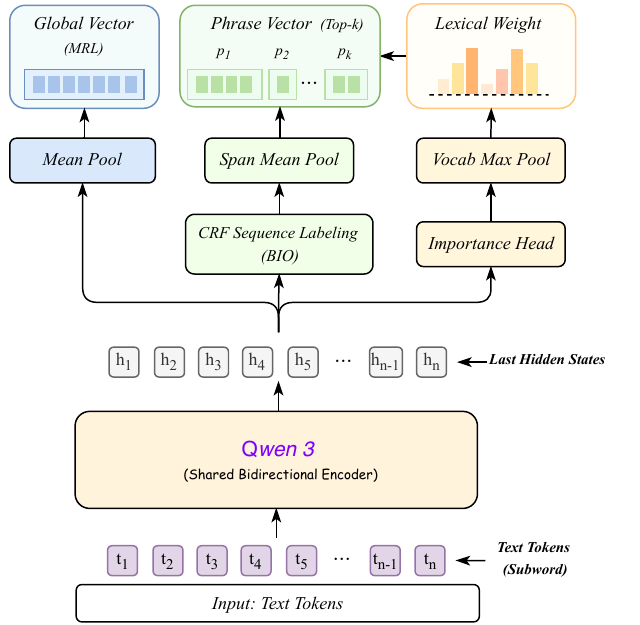}
    \caption{Architecture of \model{}. Token importance scores are max-pooled by vocabulary into lexical weights and aggregated within CRF-defined phrases into phrase weights for top-$B$ selection and query-side weighted MaxSim.}
    \label{fig:architecture}
\end{figure}

As illustrated in Figure~\ref{fig:architecture}, \model{} is a single-backbone unified multi-granularity bi-encoder built on a shared bidirectional Qwen3 encoder. Its contextual token states support global and phrase retrieval branches together with an auxiliary lexical scoring view: a global vector for single-vector retrieval, context-dependent phrase vectors for budgeted multi-vector retrieval, and lexical weights for exact-match scoring. The phrase branch provides an intermediate retrieval granularity between global compression and token-level interaction. The following sections present the model components, training procedure, and inference cost.

\subsection{Shared Bidirectional Encoder}
\label{sec:shared_encoder}

\model{} uses a shared-backbone bi-encoder: queries and documents are
encoded independently by the same bidirectional Qwen3 encoder. We
initialize Qwen3-0.6B-Base and replace its causal attention mask with a
padding-aware bidirectional mask. Given a token sequence
$x=(x_1,\ldots,x_n)$, the encoder produces contextual token states
\begin{equation}
    \mathbf{H}_x=f_\theta(x)
    =[\vh_1,\ldots,\vh_n],
    \qquad \vh_i\in\mathbb{R}^{d}.
    \label{eq:shared_encoder}
\end{equation}

Here, $d=1024$ is the encoder hidden size, while $r\leq d$ denotes the retrieval representation width. We write $\vh_i^{(r)}$ for the first $r$ dimensions of $\vh_i$, using $r=1024$ for the full-dimensional model and $r=128$ for the compact MRL-128. All retrieval views are derived from the shared states $\mathbf{H}_x$.

\subsection{Learning Context-Dependent Retrieval Units}
\label{sec:phrase_partition}

A linear-chain CRF \citep{lafferty2001crf} predicts a BIO label $y_i\in\{O,B,I\}$ for each contextual token state. Each $B$ tag starts a phrase, and subsequent $I$ tags extend it. Tokens labeled $O$ are retained as singleton units. The decoded units form a partition $\mathcal{S}(x)=\{s_1,\ldots,s_m\}$ of the token positions, so every token belongs to one retrieval unit. Retaining $O$ tokens as singletons ensures that uncertain boundary predictions do not discard local retrieval signals.

For each unit $s\in\mathcal{S}(x)$, we mean-pool its contextual states into a vector and then apply $\ell_2$ normalization:
\begin{equation}
    \vp_{x,s}^{(r)}
    =
    \operatorname{norm}\!\left(
        \frac{1}{|s|}
        \sum_{i\in s}\vh_i^{(r)}
    \right).
    \label{eq:phrase_pool}
\end{equation}
Because the CRF predicts boundaries from bidirectional contextual states, the same surface sequence may even form different retrieval units in different linguistic contexts.

An LLM teacher provides explicit BIO supervision for the CRF boundary predictor, covering domain terms, abbreviations, compounds, numerical expressions, and punctuation. An optional domain-specific terminology trie may further apply deterministic boundary overrides at inference time.

\subsection{Budgeted Unit Selection and Weighted Phrase Interaction}
\label{sec:budgeted_phrase}

Partitioning determines which retrieval units are available, while importance-guided selection determines which units are retained under the local-vector budget. Importance scores control how strongly query units contribute during phrase interaction. A shared importance head assigns each token a positive score, which is summed within each unit:

\begin{equation}
u_{x,i}=\operatorname{softplus}\!\left(
\mathbf{w}_{\mathrm{imp}}^\top\vh_i+b_{\mathrm{imp}}
\right),
\;
a_x(s)=\sum_{i\in s}u_{x,i}.
\label{eq:unit_importance}
\end{equation}
Softplus yields positive, unbounded scores, $u_{x,i}\in(0,\infty)$; they represent relative importance rather than probabilities.

Because $\mathcal{S}(x)$ is by construction a complete partition,
\begin{equation}
\sum_{s\in\mathcal{S}(x)} a_x(s)
=
\sum_{i=1}^{n} u_{x,i}.
\label{eq:importance_preservation}
\end{equation}
Thus, grouping tokens changes the interaction granularity while preserving their total importance before truncation.

Given a local-vector budget $B$, we retain up to $B$ units with the largest importance scores at inference time:
\begin{equation}
    \widehat{\mathcal{S}}_{B}(x)
    =
    \operatorname{TopB}_{s\in\mathcal{S}(x)}
    a_x(s).
    \label{eq:topb_selection}
\end{equation}

For query and document budgets $(B_q,B_d)$, we compute weighted phrase MaxSim between retained units as
\begin{equation}
    S_P^{(r)}(q,d)=
    \sum_{s\in\widehat{\mathcal{S}}_{B_q}(q)}\!
    a_q(s)\!
    \max_{t\in\widehat{\mathcal{S}}_{B_d}(d)}\!
    \bigl(\vp_{q,s}^{(r)}\bigr)^\top\!
    \vp_{d,t}^{(r)}.
    \label{eq:phrase_score}
\end{equation}
Unit importance is used for budgeted selection on both sides. On the query side, it weights each unit's MaxSim contribution; document-side importance is used only for selection. Contextual boundaries therefore determine which token states interact jointly, while importance determines which units are retained and how strongly query units contribute.

\subsection{Supporting Global and Lexical Views}
\label{sec:supporting_views}

\paragraph{Global view.}
The global view supports single-vector retrieval by summarizing each text with one normalized representation. Using the retrieval width $r$ defined above, we mean-pool the contextual states and apply $\ell_2$ normalization:
\begin{equation}
    \vg_x^{(r)}
    =
    \operatorname{norm}\!\left(
        \frac{1}{n}
        \sum_{i=1}^{n}\vh_i^{(r)}
    \right),
    \;
    S_G^{(r)}(q,d)
    =
    \left(\vg_q^{(r)}\right)^\top
    \vg_d^{(r)}.
    \label{eq:global_view}
\end{equation}
Matryoshka representation learning (MRL) \citep{kusupati2022mrl} makes prefix widths $r$ usable within the same encoder.

\paragraph{Lexical view.}
The lexical view provides exact token-ID matching. For each token ID $t$ observed in $x$, we take the maximum importance score over its occurrences:
\begin{equation}
    w_x(t)
    =
    \max_{i:\operatorname{id}(x_i)=t}
    u_{x,i},
    \;
    S_L(q,d)
    =
    \sum_t w_q(t)w_d(t).
    \label{eq:lexical_score}
\end{equation}
Max pooling prevents token-ID occurrences from disproportionately increasing the lexical score. The lexical view provides exact-match signals without vocabulary expansion. Global and Phrase may retrieve independently, while the lexical score serves as an auxiliary view over retrieved candidates and may be combined with either dense score through Eq.~\ref{eq:fusion}.

\subsection{Two-Stage Training}
\label{sec:training}

\paragraph{Stage 1: General Embedding Adaptation.}
We first train the global branch on weakly supervised query-document pairs. MRL applies the contrastive objective in parallel at representation widths $\mathcal{R}=\{64,128,256,512,1024\}$:
\begin{equation}
    \mathcal{L}_{\mathrm{stage1}}
    =
    \frac{1}{|\mathcal{R}|}
    \sum_{r\in\mathcal{R}}
    -\log
    \frac{
        \exp\!\left(S_G^{(r)}(q,d^+)/\tau\right)
    }{
        \sum_{d\in\mathcal{A}_q}
        \exp\!\left(S_G^{(r)}(q,d)/\tau\right)
    },
    \label{eq:stage1}
\end{equation}
where $d^+$ is the positive document, $\mathcal{A}_q$ contains the positive and in-batch negatives, and $\tau$ is the contrastive temperature. Only the global branch is optimized in this stage.

\paragraph{Stage 2: Joint Multi-Granularity Fine-Tuning.}
Stage~2 jointly optimizes three retrieval heads $h\in\{G,T,L\}$, corresponding to global, token-interaction, and lexical retrieval. For each query, teacher logits define a soft target over an explicit positive-negative group $\mathcal{D}_q$ during training:
\begin{equation}
\begin{aligned}
    \pi^*(d\mid q)
    &=
    \operatorname{softmax}_{d\in\mathcal{D}_q}
    t(q,d), \\
    \pi_h(d\mid q)
    &=
    \operatorname{softmax}_{d\in\mathcal{D}_q}
    \frac{S_h(q,d)}{\tau}.
\end{aligned}
\label{eq:teacher_distribution}
\end{equation}
Here, $S_h$ denotes $S_G^{(r)}$, $S_T$, or $\widetilde{S}_L$ for $h=G,T,L$, respectively. For the global head, the distribution and loss are evaluated separately at each $r\in\mathcal{R}$ and then averaged.

Each retrieval head combines hard-label contrastive learning with teacher distillation within a unified objective:
\begin{equation}
    \mathcal{L}_h
    =
    \mathcal{L}_h^{\mathrm{NCE}}
    +
    D_{\mathrm{KL}}\!\left(
        \pi^*(\cdot\mid q)
        \,\|\,
        \pi_h(\cdot\mid q)
    \right).
    \label{eq:retrieval_head_loss}
\end{equation}
For the global head, the loss is averaged over the representation widths in $\mathcal{R}$. The contrastive term uses in-batch negatives, with cross-device negatives additionally used for the global head, whereas distillation is computed over $\mathcal{D}_q$.

The phrase partition is not differentiated through the retrieval objective. Instead, the shared contextual states receive retrieval supervision through a late-interaction head. A learned projection produces normalized token vectors
\begin{equation}
    \mathbf{z}_{x,i}
    =
    \operatorname{norm}\!\left(
        \mathbf{W}_T\vh_i+\mathbf{b}_T
    \right),
    \label{eq:token_projection}
\end{equation}
with the training score defined by token-level MaxSim
\begin{equation}
    S_T(q,d)
    =
    \frac{1}{n_q}
    \sum_{i=1}^{n_q}
    \max_{1\leq j\leq n_d}
    \mathbf{z}_{q,i}^{\top}\mathbf{z}_{d,j}.
    \label{eq:token_training_score}
\end{equation}

For lexical training, repeated token IDs are summed:
\begin{equation}
\label{eq:lexical_training_score}
\widetilde{w}_x(t)=
\sum_{i:\operatorname{id}(x_i)=t}\!u_{x,i},
\quad
\widetilde{S}_L(q,d)=
\sum_t\widetilde{w}_q(t)\widetilde{w}_d(t).
\end{equation}
We obtain $\mathcal{L}_L$ during training by applying Eq.~\ref{eq:retrieval_head_loss} to $\widetilde{S}_L$.

The CRF is trained with teacher-provided BIO labels from an LLM for each query and its positive document:
\begin{equation}
    \mathcal{L}_{\mathrm{seg}}
    =
    -\log p_\phi
    \left(y_q^{\mathrm{BIO}}\mid q\right)
    -
    \log p_\phi
    \left(y_{d^+}^{\mathrm{BIO}}\mid d^+\right).
    \label{eq:seg_loss}
\end{equation}

The complete Stage-2 objective is defined as
\begin{equation}
    \mathcal{L}_{\mathrm{stage2}}
    =
    \lambda_G\mathcal{L}_G
    +
    \lambda_T\mathcal{L}_T
    +
    \lambda_L\mathcal{L}_L
    +
    \lambda_{\mathrm{seg}}\mathcal{L}_{\mathrm{seg}}.
    \label{eq:stage2_loss}
\end{equation}

Inference combines retrieval-sensitive states learned through $\mathcal{L}_T$, context-dependent boundaries learned through $\mathcal{L}_{\mathrm{seg}}$, and importance scores learned through $\mathcal{L}_L$. The token-projection head is training-only and discarded at inference.

\subsection{Inference and Indexing Cost}
\label{sec:inference_cost}

Each document is encoded once and stored in two separate dense retrieval indices: one global vector and up to $B_d$ phrase vectors. At query time, the global and phrase branches search independently using inner-product similarity and weighted MaxSim over at most $B_q$ query units, respectively. Their results may be used separately or fused with lexical scores. For representation width $r$, scoring one query-document pair with phrase interaction costs $O(B_q B_d r)$, while the global and phrase indices require $O(|\mathcal{D}|,r)$ and $O(|\mathcal{D}|,B_d,r)$ storage, respectively. End-to-end retrieval cost depends on the indexing backend and approximation strategy.

\section{Experiments}
\label{sec:experiments}

Our evaluation systematically addresses four central research questions.
\textbf{RQ1: Effectiveness.} Does phrase retrieval consistently outperform the corresponding global branch?
\textbf{RQ2: Retrieval-Unit Quality.} Under matched vector budgets, how do contextual phrases compare with token, fixed, and randomized units?
\textbf{RQ3: Mechanism.} Which design choices drive phrase retrieval performance?
\textbf{RQ4: Quality-Cost Trade-Off.} Does phrase interaction provide a practical operating point between global and token-level retrieval?

\subsection{Experimental Setup}

\paragraph{Training data and compute.}
Stage~1 uses approximately 15 million weakly supervised query-document pairs, primarily from general-domain sources including CC-pair, BAAI-MTP, and Wikipedia, with additional medical data from Huatuo-QA and PubMed. Stage~2 uses 1.51 million retrieval examples from 12 datasets, comprising approximately 88\% general-domain and 12\% medical data. Training across both stages uses 64 AMD Instinct MI308X accelerators.

\paragraph{Benchmarks and metrics.}
We evaluate on 16 MTEB/C-MTEB tasks: NFCorpus \citep{boteva2016nfcorpus}, SciFact \citep{wadden2020scifact}, MedicalQA \citep{benabacha2019medicalqa}, ChatDoctor \citep{li2023chatdoctor}, FeedbackQA \citep{li2022feedbackqa}, eight R2MED subsets \citep{zhang2025r2med}, and three C-MTEB tasks \citep{xiao2023bge,muennighoff2023mteb}. We report exponential-gain nDCG@10. Controlled mechanism, matched-candidate, and pool-size analyses use NFCorpus, SciFact, MedicalQA, and FeedbackQA. Dimensional and low-budget stress tests add CUREv1 and PublicHealthQA-en/zh, forming a separate 19-task suite.

\paragraph{Baselines and implementation.}
Dense baselines use model-native prompting and pooling with exact full-corpus search. ConstBERT-32 \citep{macavaney2025constbert} is the fixed-count, non-token-aligned baseline; ColBERTer-BOW2 \citep{hofstatter2022colberter} and LateOn-HPool \citep{clavie2024tokenpooling,chaffin2026hpoolregularization} compress token-derived representations; and ColBERTv2, BGE-M3, and Jina-ColBERT-v2 provide token-level references for comparison \citep{santhanam2022colbertv2}.

The late-interaction baselines score a union of the top-1,000 dense results from the frozen \model{} global branch, Qwen3-Embedding-0.6B, and BGE-M3. These results are neither controlled head ablations nor full-corpus multi-vector evaluations. Statistical analyses use paired query-level comparisons, 10,000 bootstrap samples, and Holm-Bonferroni correction. Latency is measured on one NVIDIA H20 GPU, and storage is measured from serialized files. See Appendices~\ref{sec:model_data_appendix}, \ref{sec:baseline_appendix}, and~\ref{sec:boundary_control_appendix} for details.

\subsection{Evaluation Protocols}
\paragraph{Retrieval settings.}

All public \model{} Phrase, Hybrid, and Lexical results in the 16-task main evaluation rerank the same full-corpus Global top-1,000 candidates; they are therefore candidate-constrained local-scoring results, not independent full-corpus phrase retrieval. The external late-interaction comparison instead reranks a frozen union of three dense top-1,000 sources, with identical candidate IDs within each reported union. Broad evaluations use CRF boundaries with terminology-trie overrides, whereas controlled studies use CRF-only boundaries. Pool-size sensitivity varies $K\in\{100,200,500,1000,2000\}$.

To separate the evaluation protocols, we use four frozen settings. \emph{Broad-deployed} denotes the 16-task CRF+trie evaluation over \modelmrl{} Global top-1,000 candidates and supplies Table~\ref{tab:main_results}, the \modelmrl{} aggregates in Appendix~\ref{sec:baseline_appendix}, and the fusion analysis. \emph{Dimensional} denotes the frozen CRF-only sweep in which each local representation scores its own same-dimensional Global top-1,000 candidates (Figure~\ref{fig:sota_four_task}; full task-level results are in Appendix~\ref{sec:baseline_appendix}). \emph{Shared-pool} denotes reranking over an identical three-source candidate union (Table~\ref{tab:shared_pool_external}).
\emph{Pool-sensitivity} varies the size of the own-Global candidate pool using natural local-unit counts (Table~\ref{tab:candidate_sensitivity}). Scores are compared only within the corresponding protocol.

\paragraph{Exact-count controls.}
Let $\mathcal{G}$ denote the grouping rules. For text $x$, every rule retains the same feasible unit count,
\begin{equation}
C_x(B)
=
\min\!\left(
B,\min_{g\in\mathcal{G}}|g(x)|
\right),
\label{eq:exact_count}
\end{equation}
using importance-based selection. We compare tokens, whole words, fixed $n$-grams, random spans, equal chunks, and CRF phrases. Boundary controls preserve phrase-length statistics while perturbing boundary locations, and document-only budgets are evaluated. The 19-task stress test varies $B\in\{4,8,16,32,64,128\}$ and the local width from 128 to 1,024. Whitespace-free Chinese tasks are reported separately for whitespace-derived whole-word grouping.

\paragraph{Human boundary audit.}
Two annotators label semantic-unit boundaries and multi-token terminology spans for 200 English and Chinese texts from eight datasets. Disagreements are adjudicated by union: a boundary is retained in the gold reference if either annotator marks it, and terminology spans from both annotators are merged. We compare CRF variants with fixed and randomized baselines using bootstrap confidence intervals and Holm-corrected paired tests.

\section{Results and Analysis}
\label{sec:analysis}

\subsection{Effectiveness (RQ1)}

\paragraph{Overall effectiveness.}
Across 16 tasks, \modelmrl{} Phrase improves over Global by 6.91 macro nDCG@10 and performs better on 15 tasks (Table~\ref{tab:main_results}). Hybrid adds only 0.33 points over Phrase, indicating that phrase interaction provides most of the multi-view gain.

\begin{table}[t]
\centering
\scriptsize
\renewcommand{\arraystretch}{0.97}

\begin{tabular*}{\columnwidth}{@{}l@{\extracolsep{\fill}}rrrr@{}}
\toprule
Task & Global & Phrase & Hybrid & Lexical \\
\midrule
NFCorpus & 31.81 & \textbf{38.48} & 36.74 & 20.18 \\
SciFact & 63.69 & 72.52 & \textbf{73.66} & 61.96 \\
MedicalQA & 68.07 & 69.74 & \textbf{71.15} & 40.07 \\
ChatDoctor$^\dagger$ & 47.68 & 51.65 & \textbf{54.70} & 40.70 \\
FeedbackQA$^\dagger$ & 63.07 & 71.96 & \textbf{72.08} & 49.56 \\
R2MED-Bioinfo. & 18.82 & 28.56 & \textbf{28.84} & 18.07 \\
R2MED-Biology & 14.64 & \textbf{37.04} & 30.94 & 15.28 \\
R2MED-IIYi & 14.93 & 18.97 & \textbf{19.07} & 12.63 \\
R2MED-MedQADiag & 3.94 & 6.67 & 6.36 & \textbf{8.36} \\
R2MED-MedXpert & 4.21 & \textbf{6.85} & 6.46 & 0.87 \\
R2MED-MedSci. & 31.44 & 34.26 & \textbf{34.82} & 20.00 \\
R2MED-PMCClinical & 12.04 & 24.99 & \textbf{27.03} & 20.47 \\
R2MED-PMCTreat. & 23.56 & 26.58 & 29.22 & \textbf{40.32} \\
MedicalRetrieval-zh$^\dagger$ & 52.20 & 54.82 & \textbf{57.25} & 45.07 \\
Covid-zh$^\dagger$ & 64.07 & \textbf{84.15} & 80.61 & 73.90 \\
Cmedqa-zh$^\dagger$ & 34.28 & 31.77 & \textbf{35.32} & 27.79 \\
\midrule
Macro (16) & 34.28 & 41.19 & \textbf{41.52} & 30.95 \\
Sensitivity (11) & 26.10 & \textbf{33.15} & 33.12 & 23.47 \\
\bottomrule
\end{tabular*}

\caption{nDCG@10 results for \modelmrl{}. Global uses full-corpus MRL-128 retrieval; Phrase, Hybrid, and Lexical score the same Global top-1,000 candidates. Phrase uses CRF+trie decoding. Bold marks the best view, and $\dagger$ denotes tasks excluded from the Sensitivity macro.}
\label{tab:main_results}
\end{table}

\paragraph{Matched-candidate comparison.}
Within a shared candidate union, Phrase nearly matches Token (62.94 vs.\ 62.95) while using 13.7\% fewer document vectors (Table~\ref{tab:shared_pool_external}). CRF-only Phrase is within 0.05 points of Token, and trie overrides change Phrase by 0.04 points. Among the compressed external checkpoints, LateOn-HPool reaches 61.77 with 40.22 vectors per document, whereas ConstBERT-32 reaches 56.00 with 32 fixed vectors. Because these systems differ in backbone and training, they characterize alternative operating points rather than isolate individual design choices.

\begin{table*}[t]
\centering
\scriptsize
\setlength{\tabcolsep}{2.5pt}
\renewcommand{\arraystretch}{0.94}

\begin{tabular*}{\textwidth}{
@{}lrrlr@{\extracolsep{\fill}}rrrrr@{}
}
\toprule
System & Params & Dim. & Local unit & Vec./doc
& NFCorpus & SciFact & MedicalQA & FeedbackQA & Macro \\
\midrule

\multicolumn{10}{@{}l}{\textit{Candidate union A}} \\
\modelmrl{} token
& 0.6B & 128 & subword & 288.59
& 37.975 & 72.987 & \textbf{69.128} & 71.718
& \textbf{62.95} \\

\modelmrl{} Phrase (CRF-only)
& 0.6B & 128 & CRF phrase & 250.90
& 38.070 & 73.337 & 68.553 & 71.647 & 62.90 \\

\modelmrl{} Phrase (deployed)
& 0.6B & 128 & CRF+trie & 249.03
& \textbf{38.223} & 73.221 & 68.598 & 71.719 & 62.94 \\

BGE-M3 ColBERT
& 0.568B & 1024 & subword & 320.86
& 34.578 & 69.553 & 67.457 & 70.568 & 60.54 \\

Jina-ColBERT-v2
& 0.568B & 128 & subword & 215.90
& 35.883 & 69.529 & 69.056 & \textbf{73.135} & 61.90 \\

\midrule
\multicolumn{10}{@{}l}{\textit{Candidate union B}} \\
ConstBERT-32
& 0.110B & 128 & fixed seq. proj. & 32.00
& 32.388 & 61.702 & 64.235 & 65.655 & 56.00 \\

ColBERTer-BOW2
& 0.066B & 32 & whole-word+prune & 79.04
& 34.607 & 49.840 & 57.722 & 60.821 & 50.75 \\

LateOn-HPool
& 0.149B & 128 & token h-pooling & 40.22
& 38.037 & 73.464 & 68.546 & 67.028 & 61.77 \\

ColBERTv2
& 0.110B & 128 & token (unpooled) & 196.64
& 34.661 & 69.182 & 65.254 & 66.393 & 58.87 \\
\bottomrule
\end{tabular*}

\caption{Matched-candidate comparison on four tasks. All systems within each union share candidate IDs. Systems are comparable only within each candidate union; Vec./doc denotes the equal-task mean number of document vectors.}
\label{tab:shared_pool_external}
\end{table*}

\begin{table}[t]
\centering
\scriptsize
\setlength{\tabcolsep}{5.0pt}
\renewcommand{\arraystretch}{1.00}

\begin{tabular}{@{}rrrrr@{}}
\toprule
$K$
& Recall@$K$
& Token
& Phrase
& \shortstack{Phrase P95\\(ms)} \\
\midrule
100   & 78.4 & 62.56 & 62.54 & 0.771 \\
200   & 81.9 & 62.65 & 62.59 & 1.214 \\
500   & 86.8 & 62.79 & 62.69 & 2.202 \\
1,000 & 91.0 & 62.93 & 62.83 & 4.379 \\
2,000 & 95.8 & 62.98 & 62.85 & 8.251 \\
\bottomrule
\end{tabular}

\caption{Pool-size sensitivity over each query's Global top-$K$ candidates on four tasks. Scores use MRL-128 natural units; latency is H20 phrase-scoring P95 per query. This sweep is distinct from the shared-union comparison in Table~\ref{tab:shared_pool_external}.}
\label{tab:candidate_sensitivity}
\end{table}

\begin{table}[t]
\centering
\scriptsize
\setlength{\tabcolsep}{1.2pt}
\renewcommand{\arraystretch}{0.94}

\begin{tabular*}{\columnwidth}{
@{}c@{\extracolsep{\fill}}rrrrrrr@{}
}
\toprule
\multicolumn{8}{@{}l}{\textbf{(a) Contextual boundary placement}} \\
$B$ & Learned & Bigram & Whole & \shortstack{Len.-hist.\\random} & \shortstack{Within-text\\shuffle} & $\Delta_{\mathrm{L-R}}$ & $\Delta_{\mathrm{L-S}}$ \\
\midrule
64 & \textbf{61.34} & 60.79 & 60.93 & 60.67 & 60.92 & $+0.67^{*}$ & $+0.42$ \\
128 & \textbf{61.87} & 61.03 & 61.31 & 61.13 & 61.21 & $+0.74^{*}$ & $+0.66^{*}$ \\
\bottomrule
\end{tabular*}

\vspace{1pt}

\begin{tabular*}{\columnwidth}{@{}l@{\extracolsep{\fill}}r@{}}
\toprule
\multicolumn{2}{@{}l}{\textbf{(b) Query weighting and selection at $B=128$}} \\
Ablation & $\Delta$ nDCG@10 [95\% CI] \\
\midrule
Phrase query: sum vs.\ uniform & $+1.66\ [1.17,2.17]$ \\
Token query: sum vs.\ uniform & $+0.12\ [-0.11,0.35]$ \\
Uniform query: Phrase vs.\ Token & $-1.34\ [-1.98,-0.73]$ \\
Phrase selection: sum vs.\ mean & $+0.27\ [0.07,0.49]$ \\
Phrase selection: sum vs.\ max & $+0.03\ [-0.06,0.11]$ \\
\bottomrule
\end{tabular*}

\caption{Mechanism ablations for \modelmrl{} on four tasks. Panel (a) compares boundary placements under matched conditions with equal per-text unit counts; $^{*}$ marks $p_{\rm Holm}<.05$ (95\% paired CIs in Appendix~\ref{sec:boundary_control_appendix}). Panel (b) reports weighting and selection contrasts at $B=128$, with positive values favoring the first setting.}
\label{tab:mechanism_analysis}
\end{table}

\begin{table}[t]
\centering
\scriptsize
\renewcommand{\arraystretch}{1.00}

\begin{tabular*}{0.62\columnwidth}
{@{}l@{\extracolsep{\fill}}r@{}}
\toprule
View & nDCG@10 \\
\midrule
Global (G)  & 56.67 \\
Phrase (P)  & 63.17 \\
Lexical (L) & 42.94 \\
G+P         & 62.94 \\
G+L         & 59.92 \\
P+L         & 63.09 \\
G+P+L       & \textbf{63.41} \\
\bottomrule
\end{tabular*}

\caption{Four-task macro nDCG@10 for the Broad-deployed views and fusion, computed from the tasks in Table~\ref{tab:main_results}. All local views score the same Global top-1,000 candidates.}
\label{tab:fusion}
\end{table}

\begin{table}[t]
\centering
\scriptsize
\setlength{\tabcolsep}{2.8pt}
\renewcommand{\arraystretch}{0.94}

\begin{tabular*}{\columnwidth}{
@{}c@{\extracolsep{\fill}}rrrr@{}
}
\toprule
\multicolumn{5}{@{}l}{\textbf{(a) Macro nDCG@10}} \\
$B$ & Token & Bigram & Random & Phrase \\
\midrule
4 & 23.47 & 25.65 & \textbf{27.21} & 25.70 \\
8 & 29.73 & 30.84 & \textbf{31.65} & 30.72 \\
16 & 33.76 & 34.30 & 34.63 & \textbf{34.86} \\
32 & 36.38 & 36.37 & 36.57 & \textbf{37.39} \\
64 & 37.45 & 37.37 & 37.27 & \textbf{38.63} \\
128 & 37.47 & 37.43 & 37.46 & \textbf{38.71} \\
\bottomrule
\end{tabular*}

\vspace{1pt}

\begin{tabular*}{\columnwidth}{
@{}c@{\extracolsep{\fill}}ccc@{}
}
\toprule
\multicolumn{4}{@{}l}{\textbf{(b) Phrase contrasts [95\% CI]}} \\
$B$ & $\Delta_{\mathrm{P-T}}$ & $\Delta_{\mathrm{P-B}}$ & $\Delta_{\mathrm{P-R}}$ \\
\midrule
4 & $+2.22$ {\scriptsize$[\!+\!1.33,+\!3.12]$} & $+0.04$ {\scriptsize$[\!-\!0.83,+\!0.89]$} & $-1.52$ {\scriptsize$[\!-\!2.42,-\!0.60]$} \\
8 & $+1.00$ {\scriptsize$[\!+\!0.19,+\!1.83]$} & $-0.11$ {\scriptsize$[\!-\!0.96,+\!0.73]$} & $-0.93$ {\scriptsize$[\!-\!1.79,-\!0.09]$} \\
16 & $+1.10$ {\scriptsize$[\!+\!0.41,+\!1.79]$} & $+0.56$ {\scriptsize$[\!-\!0.21,+\!1.34]$} & $+0.23$ {\scriptsize$[\!-\!0.53,+\!0.99]$} \\
32 & $+1.02$ {\scriptsize$[\!+\!0.43,+\!1.62]$} & $+1.02$ {\scriptsize$[\!+\!0.30,+\!1.74]$} & $+0.83$ {\scriptsize$[\!+\!0.11,+\!1.53]$} \\
64 & $+1.17$ {\scriptsize$[\!+\!0.64,+\!1.70]$} & $+1.26$ {\scriptsize$[\!+\!0.59,+\!1.95]$} & $+1.35$ {\scriptsize$[\!+\!0.71,+\!2.01]$} \\
128 & $+1.24$ {\scriptsize$[\!+\!0.73,+\!1.75]$} & $+1.27$ {\scriptsize$[\!+\!0.62,+\!1.94]$} & $+1.25$ {\scriptsize$[\!+\!0.60,+\!1.92]$} \\
\bottomrule
\end{tabular*}

\caption{Exact-count results for \modelmrl{} on 19 tasks with matched query and document unit counts. Panel (a) reports macro nDCG@10; Panel (b) reports Phrase contrasts. Phrase is CRF-only; Random denotes independently sampled random spans and is not length-matched to each text's learned partition. Brackets show 95\% paired-bootstrap CIs.}
\label{tab:budget_transfer}
\end{table}

\paragraph{View complementarity.}
On the four-task subset of the same Broad-deployed snapshot, Phrase is the strongest individual view, exceeding Global by 6.50 points (Table~\ref{tab:fusion}). Neither Global nor Lexical improves Phrase when added alone, while three-way fusion provides a modest 0.24-point gain. Thus Phrase is the primary effectiveness source within the candidate-constrained system; Global supplies the candidate gate and Global/Lexical retain modest complementarity.
\paragraph{Dimensional calibration.}
In a frozen CRF-only dimensional sweep, Phrase obtains 62.90 at both 128 and 1,024 dimensions, whereas Global improves from 55.61 to 56.70 (Figure~\ref{fig:sota_four_task}). This result is used only to assess dimensional sensitivity. Under their respective retrieval protocols, Phrase obtains 62.90 compared with 61.58 for Qwen3-Embedding-0.6B dense, although larger dense models remain stronger.

\subsection{Retrieval-Unit Quality (RQ2)}
\label{sec:rq2}
Under matched query and document unit counts, Phrase exceeds Token at every evaluated budget, including a $+2.22$ gain at the exploratory $B=4$ point (Table~\ref{tab:budget_transfer}); all five pre-specified Phrase-Token contrasts survive the planned Holm family. At $B\leq16$, Phrase and Bigram are statistically comparable, while Random spans lead at $B=4$ and $B=8$ and are indistinguishable from Phrase at $B=16$. From $B=32$ onward, Phrase is significantly stronger than both Bigram and Random while retaining its significant advantage over Token, making it the strongest of the four displayed retrieval-unit choices. At $B=128$, the respective Phrase gains over Token, Bigram, and Random are $+1.24$, $+1.27$, and $+1.25$. This crossover supports the central claim at moderate budgets: once the cap retains enough contextual units, learned phrase partitioning is more effective than either subword interaction or content-independent grouping. The advantage is budget-dependent rather than universal at extreme caps.

\subsection{Mechanism (RQ3)}
\label{sec:allocation}

Panel (a) isolates boundary-placement effects under matched representations, candidates, and per-text unit counts. Phrase outperforms both randomized placements at $B=128$ and Length-histogram random at $B=64$. Because Within-text shuffle preserves each text's phrase-length multiset and unit count, the $B=128$ contrast shows that where boundaries fall matters beyond span length or vector count alone.

Panel (b) shows that query-side importance weighting is the stronger mechanism. Replacing summed phrase weights with uniform weights reduces Phrase by 1.66 nDCG@10 points, whereas the corresponding Token effect is small and statistically inconclusive. The precise sum/mean/max rule used for top-$B$ selection has comparatively little effect. Together, these results show that contextual boundaries define stronger retrieval units, while query-side importance determines how selectively they contribute to phrase interaction.

\paragraph{Human-reference boundary audit.}
Against the adjudicated human reference, the CRF achieves a Boundary F1 of $0.900$ [$0.866$, $0.931$] over 200 items and significantly outperforms heuristic baselines, supporting the linguistic plausibility of its context-dependent units. Human-reference alignment is not correlated with phrase-token retrieval gains across 168 queries, indicating that the two evaluations capture complementary properties of the learned phrases.

\subsection{Quality-Cost Trade-Off (RQ4)}
\label{sec:efficiency}
Phrase interaction provides an operating point between global compression and token-level retrieval. Under the Shared-pool protocol, Phrase nearly matches Token ($62.94$ vs.\ $62.95$ macro nDCG@10) while using 13.7\% fewer document vectors (Table~\ref{tab:shared_pool_external}). Under Pool-sensitivity, increasing the own-Global pool from $K=1{,}000$ to $2{,}000$ improves Phrase by 0.02 points while nearly doubling scoring latency, supporting $K=1{,}000$ as the candidate-constrained setting (Table~\ref{tab:candidate_sensitivity}). In the frozen Dimensional sweep, increasing width from 128 to 1,024 improves Global by 0.86 points but leaves local interaction essentially unchanged, showing that MRL-128 already preserves local-interaction quality (Appendix~\ref{sec:full19_followup}). Finally, on 10.0 million documents, Phrase exceeds Global in nDCG@100 ($0.8777$ vs.\ $0.8617$), confirming that its effectiveness persists at large retrieval scale.

\section{Related Work}
\label{sec:related}

\paragraph{From Single-Vector Retrieval to Fine-Grained Interaction.}
Dense bi-encoders compress each query and document into a single vector, enabling efficient approximate nearest-neighbor search and increasingly strong embedding models \citep{reimers2019sbert,karpukhin2020dpr,izacard2022contriever,wang2022e5,zhang2025qwen3emb}. This compression, however, limits the relevance patterns expressible by dot-product retrieval, with recent theory showing that realizable top-$k$ result sets are constrained by embedding dimension \citep{weller2026limitations}. Token-level late-interaction models preserve finer-grained signals through contextualized MaxSim \citep{khattab2020colbert,santhanam2022colbertv2,jha2024jinacolbert}, but still treat tokenizer tokens as fixed atomic units. Our work instead learns flexible context-dependent phrase units at an intermediate granularity between global compression and token-level interaction.

\paragraph{From Phrase Retrieval to Context-Dependent Units.}
Phrase-retrieval methods index text spans as answer targets and extend phrase representations to passage- and document-level retrieval \citep{lee2021densephrases,lee2021phrase}, but phrases remain retrieved objects rather than the atomic units of document interaction. Multi-view models unify dense, sparse, and token-level representations within one encoder, yet retain predefined granularities \citep{chen2024bgem3}. In contrast, \model{} learns a context-dependent partition of each input, preserves uncovered tokens as singleton units, and uses the phrase units directly in weighted MaxSim, making retrieval granularity itself learnable.

\paragraph{From Multi-Vector Compression to Budgeted Unit Learning.}
Efficiency-oriented late-interaction methods reduce cost by aggregating or pruning token representations, pooling document vectors, or producing a fixed number of non-token-aligned vectors \citep{hofstatter2022colberter,clavie2024tokenpooling,chaffin2026hpoolregularization,macavaney2025constbert}. Matryoshka representation learning instead reduces the dimensionality of each stored vector \citep{kusupati2022mrl}. These methods primarily compress representations after their interaction units have been predefined. In contrast, \model{} jointly learns which context-dependent units should be formed and which should be retained under a per-text vector budget, while aggregated importance also governs query-side interaction.

\section{Discussion}
\label{sec:discussion}

Our findings suggest that retrieval capacity depends not only on representation width but also on interaction granularity. Wider vectors increase per-vector capacity, whereas learned units determine how relevance evidence is decomposed and matched; the dimensional sweep indicates that these axes are complementary rather than interchangeable. The boundary controls suggest that retrieval-optimal units are not reducible to tokenizer boundaries or linguistic segmentation, since boundary placement matters while human-reference agreement does not track retrieval gains. Phrase interaction can therefore be viewed as conditional allocation of matching resolution: Global selects where computation is spent, while learned boundaries and importance weights determine how it is distributed within each candidate. This perspective motivates adapting candidate depth, interaction granularity, and vector budgets rather than fixing them independently.

\section{Conclusion}

We introduced \model{}, a shared-encoder multi-granularity retriever that learns context-dependent phrase units, budgeted unit selection, and weighted phrase interaction. Controlled evaluations show that learned phrases improve over global compression, outperform fixed or randomized units under practical budgets, and retain token-level effectiveness with fewer document vectors. These findings establish context-dependent phrases as an intermediate granularity between single-vector and token-level retrieval.

\paragraph{Limitation.}
Current model learns phrase boundaries from teacher-provided BIO labels and uses fixed per-text vector budgets. Future work will explore retrieval-driven phrase induction, adaptive budgeting, and full-corpus phrase indexing.

\bibliography{aaai2027}
\clearpage
\appendix
\captionsetup{hypcap=false}

\section*{Appendix Overview}
This appendix provides additional training and reproducibility details, complete external-baseline results, and extended controlled analyses accompanying the main text. It further reports full-suite dimensional and vector-budget stress tests, boundary-placement and representation-space controls, large-scale retrieval results, and human-reference evaluation of the learned phrase boundaries. Together, these materials clarify the evaluation protocols and provide additional evidence for the effectiveness, mechanisms, and quality-cost trade-off of context-dependent phrase interaction.
\section{Training Data and Reproducibility}
\label{sec:model_data_appendix}

\model{} is initialized from Qwen3-0.6B-Base as described in Section~\ref{sec:method} and is not initialized from Qwen3-Embedding. Its Stage~2 mixture includes the public component in Table~\ref{tab:open_sources} together with proprietary retrieval data.

\begin{center}
\scriptsize
\setlength{\tabcolsep}{3.2pt}
\begin{tabular}{@{}lrl@{\hspace{1.1em}}lrl@{}}
\toprule
Source & Rows & Lang. & Source & Rows & Lang. \\
\midrule
MS MARCO & 485,905 & en & en-NLI & 274,951 & en \\
med mixture & 133,028 & en/zh & mMARCO-zh & 100,000 & zh \\
T2Ranking & 90,467 & zh & SQuAD & 87,599 & en \\
HotpotQA & 84,516 & en & DuReader & 80,416 & zh \\
zh-NLI & 64,707 & zh & Natural Questions & 58,568 & en \\
cMedQAv2 & 50,000 & zh & Law-Medical & 500 & en \\
\midrule
\multicolumn{6}{@{}r@{}}{Total source rows: 1,510,657} \\
\bottomrule
\end{tabular}
\captionof{table}{Public 12-source component of the \model{} Stage~2 mixture. Language totals are 992,039 English, 385,590 Chinese, and 133,028 mixed English-Chinese rows.}
\label{tab:open_sources}
\end{center}

\paragraph{Segmentation supervision.}
The teacher is instructed to return minimal self-contained semantic units, preserve domain terms, abbreviations, compounds, case, numbers, and punctuation, and avoid rewriting the input. Returned segments are aligned deterministically to Qwen3 tokens. On a 964,155-row diagnostic subset with complete teacher-score and BIO-label coverage (omitting large MS MARCO, Natural Questions, and zh-NLI shards), there are 1,739,921 multi-token teacher segments. The typical multi-token span has median length two tokens, and source-level means range approximately 2.4-2.9; these statistics characterize the observed label distribution rather than the full mixture. The archived labels do not retain the exact teacher checkpoint identifier originally used. The CRF is trained on the BIO targets, and the trie is an inference-only deterministic override.

\section{Additional Final-Model Results}
\label{sec:final_model_appendix}

\begin{figure}[t]
\centering
\includegraphics[width=0.90\columnwidth]{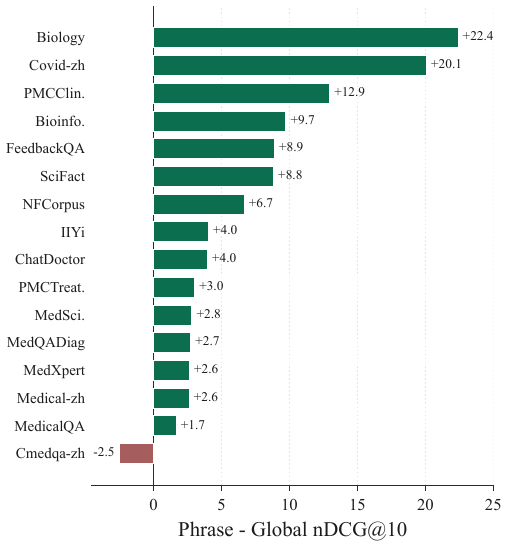}
\caption{Task-level deployed Phrase-Global nDCG@10 differences for \modelmrl{} across all evaluation tasks.}
\label{fig:model_b_gains}
\end{figure}

Figure~\ref{fig:model_b_gains} reports task-level deployed Phrase-Global nDCG@10 differences for \modelmrl{}; the corresponding branch scores appear in Table~\ref{tab:main_results}. Fifteen of sixteen differences are positive (Cmedqa is the exception).

\section{External Retrieval Baselines}
\label{sec:baseline_appendix}

Table~\ref{tab:external_validated} reports complete 16-task aggregates under the Broad-deployed protocol of Table~\ref{tab:main_results}, together with external dense and multi-vector baselines. Each aggregate is independently recomputed from its 16 task records: a genuine zero counts as completed, and missing tasks are not imputed. All displayed external configurations complete 16/16 tasks. External dense rows use exact full-corpus search. The two \modelmrl{} rows reuse the Broad-deployed MRL-128 Global and CRF+trie Phrase scores from Table~\ref{tab:main_results}. Separately, Table~\ref{tab:external_common4} and Table~\ref{tab:external_tasks} expand the CRF-only dimensional sweep summarized in Figure~\ref{fig:sota_four_task}, pairing \modelmrl{} (128 dimensions) with \model{} (1,024 dimensions); those rows are not the Broad-deployed snapshot and are not averaged to recover Table~\ref{tab:main_results}. BGE-M3 and Jina-ColBERT-v2 rerank their own dense top-200 pools, whereas the four added checkpoints rerank one shared three-source union; neither setting is an equal-cost dense baseline.

The shared-union panel in Tables~\ref{tab:external_validated} and~\ref{tab:external_tasks} is heterogeneous: ConstBERT is a fixed-count, non-token-aligned projection; ColBERTer and LateOn are token-derived count-compression methods; and ColBERTv2 is an unpooled token-level reference. Table~\ref{tab:shared_pool_external} reports their common-four geometry and vector counts, and Table~\ref{tab:shared_pool_full17} extends the comparison to 17 tasks. Protocol labels clearly distinguish union-pool reranking from full-corpus or independent-index retrieval.

\begin{center}
\smallskip
\scriptsize
\setlength{\tabcolsep}{2.4pt}
\renewcommand{\arraystretch}{0.92}
\resizebox{\columnwidth}{!}{%
\begin{tabular}{@{}lrrlrrrrr@{}}
\toprule
System & Params & Dim. & Retr. & NFC & SciFact & MedQA & FeedbackQA & Macro \\
\midrule
\modelmrl{} Global & 0.6B & 128 & single & 30.73 & 60.66 & 68.31 & 62.75 & 55.61 \\
\model{} Global & 0.6B & 1024 & single & 31.82 & 63.68 & 68.14 & 63.17 & 56.70 \\
\modelmrl{} Phrase & 0.6B & 128 & phrase & 38.10 & 73.22 & 68.51 & 71.76 & 62.90 \\
\model{} Phrase & 0.6B & 1024 & phrase & 37.65 & 73.93 & 67.94 & 72.07 & 62.90 \\
\midrule
Qwen3-0.6B & 0.6B & 1024 & dense & 35.77 & 69.92 & 71.21 & 69.40 & 61.58 \\
Qwen3-0.6B MRL & 0.6B & 128 & dense & 30.29 & 62.43 & 67.77 & 62.87 & 55.84 \\
Qwen3-4B & 4B & 2560 & dense & 41.04 & 77.46 & 75.75 & 73.08 & 66.83 \\
Qwen3-8B & 8B & 4096 & dense & \textbf{41.51} & \textbf{78.62} & \textbf{79.30} & \textbf{74.37} & \textbf{68.45} \\
BGE-M3 & 0.568B & 1024 & dense & 31.48 & 64.36 & 68.17 & 69.45 & 58.37 \\
\bottomrule
\end{tabular}%
}
\captionof{table}{Four-task CRF-only dimensional calibration with larger dense checkpoints, including per-task and macro comparisons (companion to Figure~\ref{fig:sota_four_task}). MRL is 128-d; unsuffixed \model{} is 1,024-d. Phrase reranks its own same-d Global top-1,000; dense rows use full-corpus search. Not interchangeable with Table~\ref{tab:fusion}; task-level expansion is Table~\ref{tab:external_tasks}.}
\label{tab:external_common4}
\end{center}

\paragraph{Native and shared-pool multi-vector baselines.}
BGE-M3 and Jina-ColBERT-v2 each rerank their own dense top-200 pool and are therefore neither matched-candidate nor end-to-end full-corpus multi-vector rankings. Jina obtains 60.68 on the common four tasks and 30.69 across all 16, compared with 60.46/35.92 for BGE-M3. Task heterogeneity shows that an own-pool macro does not reliably isolate local interaction quality alone. Reported Jina results use the official remote-code mapping with PyLate query/document markers and verified cross-process determinism (Table~\ref{tab:external_validated}; shared-pool comparison in Table~\ref{tab:shared_pool_external}). Table~\ref{tab:shared_pool_details} reports candidate-count diagnostics for the original union. For completeness, the four-model extension uses the same three frozen pool sources and top-1,000 recipe with a separately materialized union cache; all four systems share identical candidate IDs. Their 17-task equal-weight macros under the shared-union protocol are 30.42 (LateOn), 23.65 (ColBERTv2), 22.41 (ConstBERT-32), and 15.22 (ColBERTer-BOW2). These scores summarize released-checkpoint reranking within this evaluation protocol rather than a homogeneous model class: only ConstBERT speaks to fixed-count, non-token-aligned projection.

\paragraph{Adapter validation.}
We reload all 13 reported dense and multi-vector configurations and score a fixed relevance probe under identical inference conditions: every configuration ranks a relevant metformin passage above an unrelated Eiffel Tower passage. Table~\ref{tab:adapter_validation} shows representative geometries and fingerprints; the full check covers every Qwen native/MRL size, EmbeddingGemma native/MRL, multilingual E5/GTE, BGE-M3 dense/ColBERT, and Jina-ColBERT. The probe detects gross prompt, pooling, checkpoint, and initialization failures and is not a retrieval-effectiveness evaluation.

\begin{center}
\scriptsize
\setlength{\tabcolsep}{2.0pt}
\renewcommand{\arraystretch}{0.88}
\resizebox{\columnwidth}{!}{%
\begin{tabular}{@{}lrrrrrrrr@{}}
\toprule
Task & Cand. & Rec. & Ora. & Tok. & CRF & Trie & BGE & Jina \\
\midrule
NFCorpus & 1728.1 & 80.3 & 95.2 & 353.6 & 301.4 & 299.0 & 389.4 & 257.0 \\
SciFact & 1760.3 & 99.7 & 99.7 & 327.4 & 266.8 & 264.0 & 374.2 & 257.6 \\
MedicalQA & 1424.2 & 99.7 & 99.7 & 231.8 & 206.4 & 204.8 & 263.2 & 161.1 \\
FeedbackQA & 1400.3 & 100.0 & 100.0 & 241.7 & 228.9 & 228.4 & 256.6 & 187.9 \\
\midrule
Mean & 1578.2 & 94.9 & 98.6 & 288.6 & 250.9 & 249.0 & 320.9 & 215.9 \\
\bottomrule
\end{tabular}%
}
\captionof{table}{Shared-pool candidate diagnostics averaged across the four tasks. Cand./Rec./Ora.: candidates/query, recall (\%), oracle nDCG@10. Tok./CRF/Trie/BGE/Jina: mean local vectors/doc. FP16 payloads 73.9/64.2/63.8/657.1/55.3\,KB/doc.}
\label{tab:shared_pool_details}
\end{center}

\begin{center}
\scriptsize
\setlength{\tabcolsep}{3.0pt}
\renewcommand{\arraystretch}{0.88}
\begin{tabular}{@{}lp{1.4cm}rrr@{}}
\toprule
Adapter & Scoring & Dim. & Pos. & Neg. \\
\midrule
Qwen3-0.6B & last tok. & 1024 & 0.725 & 0.104 \\
BGE-M3 dense & mean & 1024 & 0.644 & 0.247 \\
BGE-M3 ColBERT & MaxSim & 1024 & 7.845 & 4.193 \\
Jina-ColBERT-v2 & MaxSim & 128 & 21.306 & 16.338 \\
\bottomrule
\end{tabular}
\captionof{table}{Adapter validation probe using identical positive-negative inputs (representative rows; all 13 configurations pass). Scores are adapter-native. Stack: PyTorch~2.8.0, Transformers~4.51.1, ST~5.3.0, PyLate~1.6.0, CUDA~12.8.}
\label{tab:adapter_validation}
\end{center}

\begin{table*}[t]
\centering
\scriptsize
\setlength{\tabcolsep}{3.1pt}
\renewcommand{\arraystretch}{0.88}
\begin{tabular}{lrrlrrrr}
\toprule
System & Params & Dim. & Retrieval & Macro (16) & Sens. (11) & EN (13) & ZH (3) \\
\midrule
\modelmrl{} global & 0.6B & 128 & single-vector & 34.28 & 26.10 & 30.61 & 50.18 \\
\modelmrl{} Phrase & 0.6B & 128 & phrase multi-vector & 41.19 & 33.15 & 37.56 & 56.91 \\
\midrule
Qwen3-Emb.-0.6B & 0.6B & 1024 & dense & 41.45 & 32.17 & 37.37 & 59.15 \\
Qwen3-Emb.-0.6B MRL & 0.6B & 128 & dense & 36.41 & 27.30 & 32.05 & 55.28 \\
Qwen3-Emb.-4B & 4B & 2560 & dense & 48.02 & 39.39 & 44.44 & 63.53 \\
Qwen3-Emb.-4B MRL & 4B & 128 & dense & 39.84 & 30.76 & 35.69 & 57.82 \\
Qwen3-Emb.-8B & 8B & 4096 & dense & \textbf{50.70} & \textbf{42.74} & \textbf{47.51} & \textbf{64.54} \\
Qwen3-Emb.-8B MRL & 8B & 128 & dense & 43.10 & 34.74 & 39.54 & 58.51 \\
BGE-M3 & 0.568B & 1024 & dense & 36.02 & 25.76 & 31.59 & 55.18 \\
EmbeddingGemma & 0.303B & 768 & dense & 17.45 & 14.33 & 16.57 & 21.26 \\
EmbeddingGemma MRL & 0.303B & 128 & dense & 10.61 & 9.02 & 10.58 & 10.77 \\
mE5-large-instruct & 0.560B & 1024 & dense & 36.66 & 26.66 & 32.59 & 54.27 \\
mGTE-base & 0.305B & 768 & dense & 41.40 & 31.62 & 36.63 & 62.07 \\
\midrule
BGE-M3 ColBERT$^\ddagger$ & 0.568B & 1024 & own dense top-200 & 35.92 & 26.09 & 31.56 & 54.86 \\
Jina-ColBERT-v2$^\ddagger$ & 0.568B & 128 & own dense top-200 & 30.69 & 23.24 & 28.23 & 41.35 \\
\midrule
ConstBERT-32$^{\mathsf U}$ & 0.110B & 128 & shared union top-1K & 20.77 & 19.82 & 23.67 & 8.19 \\
ColBERTer-BOW2$^{\mathsf U}$ & 0.066B & 32 & shared union top-1K & 13.17 & 13.41 & 16.09 & 0.49 \\
LateOn-HPool$^{\mathsf U}$ & 0.149B & 128 & shared union top-1K & 29.02 & 23.49 & 28.17 & 32.68 \\
ColBERTv2$^{\mathsf U}$ & 0.110B & 128 & shared union top-1K & 21.94 & 21.58 & 25.17 & 7.95 \\
\bottomrule
\end{tabular}
\caption{Complete 16-task nDCG@10 aggregates across dense, native multi-vector, and shared-union retrieval settings with protocol-specific annotations. \modelmrl{} rows are Broad-deployed MRL-128 from Table~\ref{tab:main_results}. Sensitivity excludes five provenance-flagged tasks; EN/ZH split by language. $\ddagger$: own dense top-200; $^{\mathsf U}$: shared union of \modelmrl{} Global, Qwen3-Embedding-0.6B, and BGE-M3 top-1,000. Only ConstBERT is fixed-count and non-token-aligned.}
\label{tab:external_validated}
\end{table*}

\begin{table*}[t]
\centering
{\fontsize{6.5}{7.2}\selectfont
\setlength{\tabcolsep}{1.6pt}
\renewcommand{\arraystretch}{0.78}
\begin{tabular}{@{}lrrl *{8}{r}@{}}
\toprule
System & Params & Dim. & Protocol & NFC. & SciFact & MedQA & ChatDr. & Feedback & Bioinfo. & Biology & IIYi \\
\midrule
\modelmrl{} global & 0.6B & 128 & single-vector & 30.73 & 60.66 & 68.31 & 45.55 & 62.75 & 20.80 & 13.35 & 12.29 \\
\model{} global & 0.6B & 1024 & single-vector & 31.82 & 63.68 & 68.14 & 47.68 & 63.17 & 19.28 & 14.58 & 14.85 \\
\modelmrl{} Phrase & 0.6B & 128 & own gate, CRF & 38.10 & 73.22 & 68.51 & 51.77 & 71.76 & 29.51 & 36.38 & 18.92 \\
\model{} Phrase & 0.6B & 1024 & own gate, CRF & 37.65 & 73.93 & 67.94 & 51.84 & 72.07 & 29.27 & 35.81 & 18.92 \\
\midrule
Qwen3-Emb.-0.6B & 0.6B & 1024 & full-corpus dense & 35.77 & 69.92 & 71.21 & 62.54 & 69.40 & 35.50 & 13.24 & 22.54 \\
Qwen3-Emb.-0.6B MRL & 0.6B & 128 & full-corpus dense & 30.29 & 62.43 & 67.77 & 53.50 & 62.87 & 31.38 & 11.15 & 19.73 \\
Qwen3-Emb.-4B & 4B & 2560 & full-corpus dense & 41.04 & 77.46 & 75.75 & 71.37 & 73.08 & 47.94 & 15.47 & 24.31 \\
Qwen3-Emb.-4B MRL & 4B & 128 & full-corpus dense & 31.90 & 67.05 & 71.99 & 60.39 & 65.14 & 36.89 & 12.56 & 19.04 \\
Qwen3-Emb.-8B & 8B & 4096 & full-corpus dense & 41.51 & 78.62 & 79.30 & 73.10 & 74.37 & 49.98 & 17.22 & 27.85 \\
Qwen3-Emb.-8B MRL & 8B & 128 & full-corpus dense & 33.93 & 71.68 & 76.83 & 64.28 & 67.69 & 43.78 & 13.26 & 22.33 \\
BGE-M3 & 0.568B & 1024 & full-corpus dense & 31.48 & 64.36 & 68.17 & 57.96 & 69.45 & 30.39 & 9.41 & 17.73 \\
EmbeddingGemma & 0.303B & 768 & full-corpus dense & 19.51 & 45.01 & 34.56 & 19.12 & 38.67 & 13.44 & 8.50 & 6.69 \\
EmbeddingGemma MRL & 0.303B & 128 & full-corpus dense & 9.64 & 26.64 & 22.26 & 12.70 & 25.57 & 12.35 & 4.73 & 6.03 \\
mE5-large-instruct & 0.560B & 1024 & full-corpus dense & 35.03 & 71.52 & 65.33 & 63.24 & 67.21 & 26.60 & 6.58 & 13.90 \\
mGTE-base & 0.305B & 768 & full-corpus dense & 36.70 & 73.42 & 66.80 & 63.82 & 64.57 & 31.72 & 12.90 & 17.88 \\
\midrule
BGE-M3 ColBERT$^\ddagger$ & 0.568B & 1024 & own dense top-200 & 34.27 & 69.48 & 67.50 & 52.68 & 70.57 & 25.44 & 11.71 & 14.47 \\
Jina-ColBERT-v2$^\ddagger$ & 0.568B & 128 & own dense top-200 & 33.89 & 67.23 & 68.59 & 38.35 & 73.01 & 21.22 & 13.85 & 8.72 \\
\midrule
ConstBERT-32$^{\mathsf U}$ & 0.110B & 128 & shared union top-1K & 32.39 & 61.70 & 64.23 & 24.02 & 65.66 & 17.21 & 6.54 & 7.55 \\
ColBERTer-BOW2$^{\mathsf U}$ & 0.066B & 32 & shared union top-1K & 34.61 & 49.84 & 57.72 & 0.79 & 60.82 & 1.04 & 2.43 & 0.19 \\
LateOn-HPool$^{\mathsf U}$ & 0.149B & 128 & shared union top-1K & 38.04 & 73.46 & 68.55 & 40.80 & 67.03 & 21.44 & 11.18 & 7.84 \\
ColBERTv2$^{\mathsf U}$ & 0.110B & 128 & shared union top-1K & 34.66 & 69.18 & 65.25 & 23.49 & 66.39 & 18.16 & 11.35 & 7.14 \\
\bottomrule
\end{tabular}}
\caption{Task-level nDCG@10 across eight displayed tasks, panel (a): CRF-only dimensional sweep (Figure~\ref{fig:sota_four_task}, Table~\ref{tab:external_common4}), not Table~\ref{tab:main_results}. $\ddagger$: own dense top-200; $^{\mathsf U}$: shared union. All systems are evaluated under their explicitly stated candidate-generation protocols. Rows jointly cover dense, phrase, native multi-vector, and shared-union configurations. Panel (b) continues below.}
\label{tab:external_tasks}

\vspace{4pt}
\ContinuedFloat
\centering
{\fontsize{6.5}{7.2}\selectfont
\setlength{\tabcolsep}{1.6pt}
\renewcommand{\arraystretch}{0.78}
\begin{tabular}{@{}lrrl *{8}{r}@{}}
\toprule
System & Params & Dim. & Protocol & MedDiag & MedX & MedSci. & PMCClin. & PMCTreat. & Medical-zh & Covid-zh & Cmedqa-zh \\
\midrule
\modelmrl{} global & 0.6B & 128 & single-vector & 4.85 & 4.77 & 29.32 & 12.13 & 27.95 & 50.74 & 64.44 & 33.49 \\
\model{} global & 0.6B & 1024 & single-vector & 4.42 & 4.21 & 31.95 & 12.03 & 28.87 & 52.16 & 63.91 & 34.22 \\
\modelmrl{} Phrase & 0.6B & 128 & own gate, CRF & 7.25 & 6.34 & 34.99 & 25.27 & 32.64 & 53.53 & 78.30 & 31.32 \\
\model{} Phrase & 0.6B & 1024 & own gate, CRF & 7.31 & 6.64 & 35.39 & 24.85 & 32.57 & 53.44 & 77.79 & 31.30 \\
\midrule
Qwen3-Emb.-0.6B & 0.6B & 1024 & full-corpus dense & 8.72 & 5.57 & 33.89 & 22.52 & 34.96 & 55.76 & 83.08 & 38.61 \\
Qwen3-Emb.-0.6B MRL & 0.6B & 128 & full-corpus dense & 6.18 & 4.20 & 28.32 & 13.57 & 25.32 & 51.45 & 78.90 & 35.49 \\
Qwen3-Emb.-4B & 4B & 2560 & full-corpus dense & 19.74 & 13.56 & 40.84 & 36.08 & 41.07 & 62.19 & 84.41 & 43.98 \\
Qwen3-Emb.-4B MRL & 4B & 128 & full-corpus dense & 10.21 & 7.64 & 33.77 & 21.73 & 25.61 & 55.89 & 78.93 & 38.65 \\
Qwen3-Emb.-8B & 8B & 4096 & full-corpus dense & 22.83 & 18.08 & 42.84 & 41.89 & 50.01 & 63.68 & 85.18 & 44.75 \\
Qwen3-Emb.-8B MRL & 8B & 128 & full-corpus dense & 16.49 & 7.03 & 34.44 & 30.34 & 32.00 & 57.18 & 79.05 & 39.31 \\
BGE-M3 & 0.568B & 1024 & full-corpus dense & 3.67 & 1.97 & 26.97 & 11.34 & 17.83 & 54.19 & 77.56 & 33.78 \\
EmbeddingGemma & 0.303B & 768 & full-corpus dense & 0.00 & 0.76 & 26.46 & 1.40 & 1.26 & 20.66 & 26.45 & 16.67 \\
EmbeddingGemma MRL & 0.303B & 128 & full-corpus dense & 0.00 & 0.00 & 16.18 & 0.55 & 0.86 & 11.38 & 8.64 & 12.28 \\
mE5-large-instruct & 0.560B & 1024 & full-corpus dense & 4.36 & 1.64 & 33.48 & 14.57 & 20.25 & 55.76 & 73.06 & 33.98 \\
mGTE-base & 0.305B & 768 & full-corpus dense & 9.72 & 6.55 & 35.77 & 17.50 & 38.85 & 61.85 & 80.61 & 43.76 \\
\midrule
BGE-M3 ColBERT$^\ddagger$ & 0.568B & 1024 & own dense top-200 & 2.78 & 2.85 & 28.89 & 15.84 & 13.74 & 52.86 & 80.79 & 30.92 \\
Jina-ColBERT-v2$^\ddagger$ & 0.568B & 128 & own dense top-200 & 0.72 & 1.14 & 33.62 & 5.11 & 1.56 & 37.08 & 67.00 & 19.96 \\
\midrule
ConstBERT-32$^{\mathsf U}$ & 0.110B & 128 & shared union top-1K & 0.16 & 1.18 & 23.29 & 3.77 & 0.00 & 7.22 & 13.86 & 3.48 \\
ColBERTer-BOW2$^{\mathsf U}$ & 0.066B & 32 & shared union top-1K & 0.19 & 0.00 & 0.00 & 0.53 & 1.00 & 0.60 & 0.36 & 0.51 \\
LateOn-HPool$^{\mathsf U}$ & 0.149B & 128 & shared union top-1K & 1.32 & 0.74 & 30.09 & 3.57 & 2.20 & 30.07 & 54.33 & 13.65 \\
ColBERTv2$^{\mathsf U}$ & 0.110B & 128 & shared union top-1K & 0.36 & 0.94 & 27.09 & 3.10 & 0.14 & 6.89 & 12.81 & 4.14 \\
\bottomrule
\end{tabular}}
\caption{Continued, panel (b): remaining eight tasks.}
\label{tab:external_tasks_b}
\end{table*}

\section{Full-Suite Dimension and Low-Budget Stress Tests}
\label{sec:full19_followup}

The primary controlled tables use four pre-specified tasks. We additionally run the same raw-state, CRF-only protocol on all 19 available tasks: the 16-task main suite plus CUREv1 and PublicHealthQA-en/zh. The 128-dimensional operating point is the MRL prefix; unsuffixed \model{} denotes the full 1,024-dimensional representation. These extensions use natural local-unit counts for the dimensional sweep and exact feasible per-text counts for the budget sweep. Every macro gives each task equal weight under both settings.

\paragraph{Local dimensionality.}
Table~\ref{tab:dimension_full19} changes both the global retrieval document set and the local prefix (``own gate''). The global representation benefits consistently when moving from MRL-128 to full 1,024 dimensions. In contrast, \model{} Token and Phrase are unchanged. When candidate IDs are frozen to the MRL-128 gate, the local full-minus-MRL effects are $-0.04/-0.12$ for \model{} Token/Phrase; none survives Holm correction. This separates global candidate capacity from local interaction capacity.

\begin{center}
\scriptsize
\setlength{\tabcolsep}{2.6pt}
\renewcommand{\arraystretch}{0.90}
\begin{tabular}{@{}llrrrl@{}}
\toprule
Model & Own-gate & MRL-128 & Full & $\Delta$ [95\% CI] & $p_{\rm H}$ \\
\midrule
\model{} & Global & 39.21 & 40.07 & $+0.86$ [$+0.51$,$+1.20$] & $<.001$ \\
\model{} & Token  & 46.35 & 46.48 & $+0.13$ [$-0.14$,$+0.39$] & 1.000 \\
\model{} & Phrase & 46.41 & 46.39 & $-0.02$ [$-0.29$,$+0.26$] & 1.000 \\
\bottomrule
\end{tabular}
\captionof{table}{Nineteen-task MRL-128-versus-full-1,024 dimensional sweep with uncertainty (macro nDCG@10). Own same-d global gate per row; paired task-stratified bootstrap CIs. Holm family: three own-gate contrasts plus two frozen-gate local contrasts reported below in the text.}
\label{tab:dimension_full19}
\end{center}

\paragraph{Aggressive exact counts.}
Table~\ref{tab:lowbudget_full19} displays the joint query/document sweep at $B\in\{4,8,16,32,64,128\}$. The pre-specified main grid is $B\in\{8,16,32,64,128\}$; $B=4$ is reported as an additional operating point. Phrase exceeds Token at every displayed budget for \modelmrl{}. All five pre-specified Phrase-Token contrasts survive the within-model ten-member Phrase-\{Token, Bigram\} Holm family. The \modelmrl{} Phrase-Bigram contrast becomes family-significant from $B=32$. Whole word and Equal chunks remain stronger in the all-task macro, so the result does not support universal phrase dominance.

\begin{center}
\scriptsize
\setlength{\tabcolsep}{2.4pt}
\begin{tabular}{@{}lrrrrrr@{}}
\toprule
Unit & $B{=}4$ & $B{=}8$ & $B{=}16$ & $B{=}32$ & $B{=}64$ & $B{=}128$ \\
\midrule
Token        & 23.47 & 29.73 & 33.76 & 36.38 & 37.45 & 37.47 \\
Whole word   & 32.89 & 36.89 & 40.00 & 41.71 & \textbf{43.17} & \textbf{43.25} \\
Fixed bigram & 25.65 & 30.84 & 34.30 & 36.37 & 37.37 & 37.43 \\
Equal chunks & \textbf{39.76} & \textbf{40.16} & \textbf{41.10} & \textbf{41.82} & 42.40 & 42.79 \\
Random spans & 27.21 & 31.65 & 34.63 & 36.57 & 37.27 & 37.46 \\
Phrase       & 25.70 & 30.72 & 34.86 & 37.39 & 38.63 & 38.71 \\
\bottomrule
\end{tabular}
\captionof{table}{Joint query/document exact-count stress across six budgets for \modelmrl{} with equal-task weighting throughout (19-task macro nDCG@10). Main grid $B\in\{8,16,32,64,128\}$; $B{=}4$ extra. Bold: best rule per budget. Interpret with the non-CJK slice in Table~\ref{tab:lowbudget_en15}.}
\label{tab:lowbudget_full19}
\end{center}

\paragraph{Why the all-task ordering reverses.}
The Whole-word rule is derived from whitespace. On Cmedqa, Covid, MedicalRetrieval, and PublicHealthQA-zh, its low natural count forces the common query count to only 1.0-1.2 units at every nominal budget. Token, Bigram, and Phrase are consequently pruned to roughly one query unit, whereas Whole word and Equal chunks behave like near-whole-query pooled representations. This diagnoses a limitation of the whitespace-based multilingual control rather than a language-invariant comparison of semantic boundary quality very broadly.

Table~\ref{tab:lowbudget_en15} therefore reports a descriptive non-CJK slice. At $B=4$-8, Equal chunks is strongest. Phrase becomes the strongest displayed rule at $B\geq32$ for \modelmrl{}. A document-only sweep, where queries retain method-specific natural units, gives the same transition consistently on these 15 tasks under this less restrictive protocol. Across all 19 document-only tasks, Whole word remains strongest, confirming that a language-aware word control is required rather than simply changing which side is budgeted.

\begin{center}
\scriptsize
\setlength{\tabcolsep}{2.4pt}
\begin{tabular}{@{}lrrrrrr@{}}
\toprule
Unit & $B{=}4$ & $B{=}8$ & $B{=}16$ & $B{=}32$ & $B{=}64$ & $B{=}128$ \\
\midrule
Token        & 22.03 & 29.78 & 34.82 & 38.18 & 39.55 & 39.57 \\
Whole word   & 26.36 & 31.44 & 35.40 & 37.57 & 39.42 & 39.52 \\
Fixed bigram & 24.56 & 30.79 & 35.19 & 37.80 & 39.06 & 39.14 \\
Equal chunks & \textbf{35.07} & \textbf{35.60} & \textbf{36.87} & 37.79 & 38.52 & 39.02 \\
Phrase       & 24.31 & 30.33 & 35.51 & \textbf{38.72} & \textbf{40.28} & \textbf{40.38} \\
\bottomrule
\end{tabular}
\captionof{table}{Non-CJK subset (15 tasks) of the same \modelmrl{} joint exact-count sweep, reporting macro nDCG@10 across all budgets with best values clearly bolded. $B{=}4$ is extra. Diagnoses the whitespace Whole-word control; not a substitute for the four-task estimand.}
\label{tab:lowbudget_en15}
\end{center}

\section{Controlled Representation-Space Details}
\label{sec:projection_details_appendix}

Table~\ref{tab:projection_details} complements the raw-space primary comparison in Table~\ref{tab:budget_transfer} with the learned token projection. The projection is affine, so pool-then-project and project-then-pool agree up to numerical precision before normalization. Full projected vectors are 1,024-dimensional and are not treated as equal-storage rows; the table uses only the first 128 projected dimensions for a dimension-matched comparison.

\begin{center}
\scriptsize
\setlength{\tabcolsep}{2.0pt}
\resizebox{\columnwidth}{!}{%
\begin{tabular}{@{}lrrrrrr@{}}
\toprule
$B$ & Raw P-T & Proj.\ tok. & Proj.\ phr. & Proj.\ P-T & Proj.\ P-B \\
\midrule
64  & $+0.31$ [$-0.16$,0.80] & 61.50 & 61.24 & $-0.26$ [$-0.76$,0.25] & $+1.80$ [0.98,2.64] \\
128 & $+0.20$ [$-0.20$,0.61] & 62.01 & 61.81 & $-0.20$ [$-0.66$,0.25] & $+2.11$ [1.36,2.90] \\
\bottomrule
\end{tabular}%
}
\captionof{table}{MRL-128 representation-space sensitivity at two controlled budget settings for \modelmrl{}, reporting raw and projected phrase-token contrasts (same counts as Table~\ref{tab:budget_transfer}). ``Proj.'' uses the training-aligned projection truncated to 128-d. Task-stratified paired bootstrap intervals.}
\label{tab:projection_details}
\end{center}

\section{Boundary-Placement Control Details}
\label{sec:boundary_control_appendix}

Table~\ref{tab:boundary_controls_full} reports the complete planned contrast family behind Panel (a) of Table~\ref{tab:mechanism_analysis}. All methods use the same \modelmrl{} checkpoint, raw 128-dimensional states, CRF-only decoding for Learned Phrase, frozen Global top-1,000 candidates, and exactly matched realized query and document unit counts. The two random controls are averaged per query over seeds $\{13,42,2027,3407,9001\}$ for each query independently. All exact-count checks pass.

\begin{center}
\scriptsize
\setlength{\tabcolsep}{2.4pt}
\begin{tabular}{@{}clrr@{}}
\toprule
$B$ & Control & $\Delta$ [95\% CI] & $p_{\rm H}$ \\
\midrule
64  & Fixed bigram & $+0.55$ [$-0.20$, $+1.27$] & .2972 \\
64  & Whole word & $+0.41$ [$-0.19$, $+1.00$] & .2972 \\
64  & Length-hist.\ random & $+0.67$ [$+0.18$, $+1.18$] & \textbf{.0444} \\
64  & Within-text shuffle & $+0.42$ [$-0.03$, $+0.88$] & .2160 \\
\addlinespace[1pt]
128 & Fixed bigram & $+0.84$ [$+0.11$, $+1.57$] & .1160 \\
128 & Whole word & $+0.56$ [$+0.03$, $+1.10$] & .1640 \\
128 & Length-hist.\ random & $+0.74$ [$+0.32$, $+1.17$] & \textbf{.0032} \\
128 & Within-text shuffle & $+0.66$ [$+0.28$, $+1.05$] & \textbf{.0056} \\
\bottomrule
\end{tabular}
\captionof{table}{\modelmrl{} boundary-placement contrasts on four tasks, paired inference with corrected significance (Learned$-$control; equal-task macro nDCG@10). $p_{\rm H}$: Holm over eight contrasts. Bootstrap 10{,}000; raw $p$: .1486, .1788, .0074, .0720, .0232, .0410, .0004, .0008.}
\label{tab:boundary_controls_full}
\end{center}

\section{Private 10M-Document Evaluation}
\label{sec:private_10m}

The private corpus contains 10,039,519 deduplicated documents and 1,000 queries (500 Chinese and 500 English). Under the same evaluation pipeline, \model{} Global/Phrase nDCG@100 are 0.8617/0.8777, and strict-level recall is 0.0567/0.1230. Phrase nDCG@100 is 0.8611 on Chinese queries and 0.8943 on English queries. These results probe large-scale bilingual retrieval under realistic corpus-scale operating conditions; the public controlled experiments remain the reproducible scientific comparisons.

\section{Human Boundary Evaluation Details}
\label{sec:human_boundary_appendix}

Table~\ref{tab:boundary_quality} reports eight boundary constructions against Annotator~A's 200 references using the stated annotation protocol. Annotator~A provided references for all items, accepting the CRF segmentation without modification on two initially missing items. Boundary F1 is item-averaged with a 10,000-resample 95\% bootstrap interval; the remaining columns are corpus-micro diagnostics. CRF exact/multi-token span F1 values are 0.904/0.680. CRF Boundary F1 exceeds Whole word by $0.091$ [0.075, 0.108], length-matched random by $0.165$ [0.139, 0.192], Bigram by $0.347$ [0.326, 0.367], and Trigram by $0.494$ [0.470, 0.518]. All heuristic comparisons, including Equal chunks, survive the stated Holm family. CRF exceeds CRF+trie by only $0.005$ [0.000, 0.016], which is not significant after Holm correction. The Teacher row is not an independent teacher-student comparison: the archived teacher boundary arrays are identical to CRF.

\begin{center}
\scriptsize
\setlength{\tabcolsep}{3.0pt}
\begin{tabular}{@{}lccc@{}}
\toprule
Method & \shortstack{Boundary F1\\{[95\% CI]}} & \shortstack{Span F1\\exact/multi} & Term recall \\
\midrule
\textbf{CRF} & \textbf{0.899 [0.865, 0.930]} & \textbf{0.9040/0.6798} & \textbf{0.7760} \\
CRF+trie & 0.8936 [0.858, 0.926] & 0.9022/0.6742 & 0.7760 \\
Teacher$^\dagger$ & 0.899 [0.865, 0.930] & 0.9040/0.6798 & 0.7760 \\
\midrule
Whole word & 0.8075 [0.774, 0.839] & 0.7705/0.0000 & 0.0000 \\
Length-matched random & 0.7338 [0.696, 0.770] & 0.6591/0.0917 & 0.1520 \\
Bigram & 0.5521 [0.527, 0.576] & 0.0788/0.1072 & 0.1800 \\
Trigram & 0.4049 [0.383, 0.427] & 0.0482/0.0501 & 0.3560 \\
Equal chunks & 0.2936 [0.266, 0.321] & 0.0218/0.0328 & 0.5800 \\
\bottomrule
\end{tabular}
\captionof{table}{Boundary quality across eight methods on 200 human items with corpus-level diagnostic metrics. Boundary F1: item-macro with bootstrap 95\% CI; span/term-recall: corpus-micro. Annotator~A is the operational reference. $\dagger$ Archived teacher arrays match CRF (not independent).}
\label{tab:boundary_quality}
\end{center}

Inter-annotator agreement on Layer-A segmentation is limited overall: mean Cohen's $\kappa$ is 0.166, symmetric Boundary F1 is 0.436, and exact-span F1 is 0.230 over 200 paired items. Mean boundary counts are 16.0 and 4.6 per item for the two annotators, respectively, so absolute method scores are reference-sensitive. Error analysis yields 172 flags across 70 items (multiple labels allowed): 56 over-segmentation, 56 modifier detachment, 52 numeric separation, and 8 abbreviation fragmentation. Among the 168 annotated queries matched to retrieval records, Boundary F1 is uncorrelated with retrieval gain (Spearman $\rho=0.007$, $p=.927$). The annotation study therefore speaks to linguistic plausibility; controlled retrieval evidence is in Panel (a) of Table~\ref{tab:mechanism_analysis}, where Learned CRF significantly exceeds both randomized placements at $B=128$ and Length-histogram random at $B=64$.

\section{Extended Shared-Pool Multi-Vector Results}
\label{sec:shared_pool_full17_appendix}

Table~\ref{tab:shared_pool_full17} extends the Shared-pool comparison of Table~\ref{tab:shared_pool_external} to 17 tasks (the frozen 16-task suite plus CUREv1; PublicHealthQA-en/zh were not evaluated for these checkpoints). Each query uses the same independently materialized union for all four systems. ConstBERT is the sole fixed-count, non-token-aligned model; ColBERTer and LateOn directly reduce token-derived vector counts, while ColBERTv2 retains token-level interaction. Mean union size ranges from 1,400.5 to 2,137.7 candidates per query. Pool recall ranges from 70.2\% to 100.0\%, so low scores on MedXpertQAExam and MedQADiag should be read together with their 73.88 and 76.22 oracle nDCG@10 ceilings. The table is a released-checkpoint reranking benchmark, not an equal-training, equal-count, or equal-cost comparison.

\begin{center}
\scriptsize
\setlength{\tabcolsep}{2.4pt}
\begin{tabular}{@{}lrrrr@{}}
\toprule
Task & ConstBERT & ColBERTv2 & ColBERTer & LateOn \\
\midrule
CUREv1 & 48.67 & 50.94 & 48.03 & \textbf{52.86} \\
Medical-zh & 7.22 & 6.89 & 0.60 & \textbf{30.07} \\
Cmedqa-zh & 3.48 & 4.14 & 0.51 & \textbf{13.65} \\
Covid-zh & 13.86 & 12.81 & 0.36 & \textbf{54.33} \\
R2:MedX & \textbf{1.18} & 0.94 & 0.00 & 0.74 \\
R2:PMCClinical & \textbf{3.77} & 3.10 & 0.53 & 3.57 \\
R2:Biology & 6.54 & \textbf{11.35} & 2.43 & 11.18 \\
R2:MedQADiag & 0.16 & 0.36 & 0.19 & \textbf{1.32} \\
R2:Bioinformatics & 17.21 & 18.16 & 1.04 & \textbf{21.44} \\
R2:MedicalSciences & 23.29 & 27.09 & 0.00 & \textbf{30.09} \\
R2:PMCTreatment & 0.00 & 0.14 & 1.00 & \textbf{2.20} \\
R2:IIYiClinical & 7.55 & 7.14 & 0.19 & \textbf{7.84} \\
ChatDoctor & 24.02 & 23.49 & 0.79 & \textbf{40.80} \\
SciFact & 61.70 & 69.18 & 49.84 & \textbf{73.46} \\
NFCorpus & 32.39 & 34.66 & 34.61 & \textbf{38.04} \\
FeedbackQA & 65.66 & 66.39 & 60.82 & \textbf{67.03} \\
MedicalQA & 64.23 & 65.25 & 57.72 & \textbf{68.55} \\
\midrule
Equal-task macro & 22.41 & 23.65 & 15.22 & \textbf{30.42} \\
\bottomrule
\end{tabular}
\captionof{table}{17-task shared-pool nDCG@10 for four released multi-vector checkpoints on the same union (\modelmrl{} Global $+$ Qwen3-Embedding-0.6B $+$ BGE-M3 top-1,000), with equal-task macro results reported below. ConstBERT: fixed-count projection; ColBERTer/LateOn: token-derived compression; ColBERTv2: token-level. Geometry in Table~\ref{tab:shared_pool_external}. Bold: best overall per task.}
\label{tab:shared_pool_full17}
\end{center}

\end{document}